# The Translation of Circumlocution in Arabic Short Stories into English

**Dalal Waadallah Shehab**¹, **Salem Yahya Fathi**²

¹Al-Noor University college , Nineveh / Mosul, Shalalat road , Iraq, ²Department of Translation,College of Arts,University of Mosul,Nineveh / Mosul, Iraq

dalalwaadallah@alnoor.edu.iq, salem_yahya@yahoo.com

**Abstract**. This study aims at identifying and analyzing circumlocution categories and subcategories in the (SL) and their renditions into the (TL) .It is based on criteria proposed for inclusion and exclusion of circumlocution .This study is concerned with the translation of literary texts, specifically short stories , from Arabic into English . It draws on four short stories selected from Arabic famous writers and their parallel translations into English. It hypothesizes that Arabic categories of circumlocution are applicable to English categories of metadiscourse, which include textual and interpersonal items. Nida's (1964) model is adopted in this study to judge the appropriateness in translation the study shows that the translators made serious decisions while opting for various techniques such as addition, subtraction and alteration. In this sense, it investigates whether the translators have successfully and appropriately managed to render the concept of Arabic circumlocution into English or not. The main problems that led to the inappropriate translations were also identified. This study concludes that there are lots of similarities between the categories of circumlocution in Arabic and the categories of metadiscourse in English. These similarities are clear when appropriate renditions are achieved.

## 1. Introduction

The concept of translation has two meanings either the product or the process: the product in which it means the text that has been translated and the process in which it is the act of producing the translation, and known as translation. In the process of translation, the translator conveys the message (the verbal and nonverbal symbols) of (SL) text producer to the (TL) receiver in such a way that preserves the same meaning and effect of the (SL) in the (TL). It exists for facilitating both linguistic and cultural transfer. Translation is not only restricted to linguistic but also it deals with two different cultures at the same time. English and Arabic are linguistically, culturally and stylistically different. Recently, more attention has been paid to English metadiscourse. It is regarded as important as discourse. Many linguists and scholars provide a well-organized system of metadiscourse which is divided into two main categories: textual and interpersonal. Accordingly, metadiscourse acts in the text (Williams, 1981; Vande Kopple, 1985; Crismore et al., 1993).In Arabic, rhetoricians and linguists are well aware of the significance of circumlocution. However, its categories are scattered and need to be systematically arranged .Examining the Arabic literature,( البالغة Rhetoric) is categorized into three disciplines: علم البيان(eloquence), علم المعاني(semantics), and علم البديع (embellishment). Most of ancient Arab rhetoricians and linguists regard circumlocution as a branch of semantics. It is argued that circumlocution is not only a linguistic phenomenon but also has a rhetorical function. This phenomenon is used for a purpose that has its effect on the addressee. In this sense , the functions of circumlocution are persuasion, clarification of the text producer's intended meaning, creation of a close relationship with the audience .They assure that the rhetorical purposes cannot be figured out unless their context is known as well as the status of the addresser. This effect is obvious when a text containing circumlocution is compared with one that does not contain circumlocution.

Much attention has been paid to circumlocution by ancient Arab rhetoricians and linguists ( Al-Jahiz , ١٩٩٨ ; Al-Askari, ١٣١٩H ; Al-Qazwini ,١٩٨٣ ; Ibn Al-Atheer , ١٩٩٠).Similarly , modern and contemporary linguists are also interested in studying this phenomenon which is linked to modern stylistic studies ( Fadil , ١٩٩٨:١٧٦; Al-Hashimi, ١٩٩٩: ٢٠٢).

## ٢. The concept and the categories of Circumlocution

Circumlocution is one of the most ancient rhetorical devices. Ancient and modern Arab rhetoricians and linguists study it. Ibn Al-Atheer (١٩٩٠: ١٢٠), defines it as ' لفائدة زيادة اللفظ على المعنى ' ) the motivated employment of using extra words in the expression of a given meaning). Ancient rhetoricians have different methods of classifying circumlocution. The first method is suggested by Al-Suyuti (١٩٧٤: ٢٥٢) .He classifies circumlocution into بسط(stretch the sentences) and زيادة (add the particles) .The second one is suggested by Ibn al-Atheer (١٩٩٠: ١٢٠) who depends on the number of the sentences that include circumlocution. The third one is suggested by Al-Qazwini (١٩٨٣:١٧٥) who classifies circumlocution according to its rhetorical purposes. According to modernists, who are interested in investigating circumlocution (Fadil, ١٩٩٨:١٧٦; Al-Hashimi, ١٩٩٩:٢٠٢), it is strongly associated with stylistics. Stylistics is an extension to the ancient rhetoric (Fadil , ١٩٩٨, p.١٧٥) . Modernists have not added any new categories even the examples they use in their studies are the same examples used by ancient rhetoricians. Recently, most Arab researchers have dealt with the concept of circumlocution in terms of Quranic texts and Hadith Sharif (Jarrar, ٢٠٠٩ ; Eskander , ٢٠٠٣ ; Aljamas , ٢٠١٠) .

Relying on the literature review of ancient Arab rhetoricians, Eskander (٢٠٠٣:٨) made an attempt to categorize circumlocution into three headings: اطناب بياني (Eloquence circumlocution), اطناب توكيدي(Emphatic circumlocution ( and اطناب توجيهي (Directive circumlocution) . This classification is based on the functions of circumlocution. This study is going to adopt this classification for being more systematic in identifying the categories of circumlocution.

### ٢,١. (Eloquence circumlocution)(الاطناب البياني)

Eloquence is an essential and important category of circumlocution. It relies on the intertextual relations of the text to determine the functional values of circumlocution. It is used for revealing and clarifying the meaning to achieve its variant purposes. It is categorized under two headings.

#### ٢,١,١. ذكر الخاص بعد العام (Mentioning the specific item after the general one).

It means paying attention to the importance of the specific item that is mentioned after the general one. حروف العطف (coordinating particles) are used in this type of circumlocution) Abbas , ١٩٩٧: ٤٨٦). Consider the following example:

١-**_مُنيةُ النصر كغيرها من بلاد ه ل ل ا الواسعة_** تتشاءم من يوم الجمعة، وأي حادث يقع فيه البد انه كارثة اكيدة ...

(Idris, ٢٠١٩:٣٢)

- **_Like other towns_**, **_Munyat al-Nasr_** was superstitious about Friday, and any event that took place on that day was viewed as a sure catastrophe.

(Husni&Newman, ٢٠٠٨:٢٧٠)

It is clear that the original writer used (منية النصرMunyat al-Nasr) specifically before كغيرها من بلاد هلال الواسعة(Like other towns) to highlight the importance of this city.

### ٢,١,٢. الايضاح بعد الابهام (Illustration after Amphibology)

It means that the speaker mysteriously mentions the meaning, and then clarifies it during his speech (Al-Sutyuti, ١٩٨٨:٢٧٢). Consider the following example:

لَوْ صيفاً تجد عندنا ذباب البقر - **_ذباب ضخم كمثال الخريف_**، كما نقول بلهجتنا. (Saleh, ١٩٩٧: ٣٣)

- If you were to come to us in summer, you would find the horse flies with us – **_enormous flies the size of young sheep_**, as we say.

(Davies, ١٩٦٧:٨٣)

The writer clarifies to the reader what is really meant by (ذباب البقر) since if we stopped until (ذباب البقر) the sentence would be ambiguous.

### ٢,٢. الاطناب التوكيدي(Emphatic circumlocution ).

It is one of the rhetorical devices that is used for confirming the meaning to the readers. Thus, it takes into consideration the state of the audience in a particular context such as:

١. خالي الذهن (open-minded). For example: علي خرج Ali left.
٢. متردد (uncertain)، For example، ان علي خرج/ Ali **did** leave.
٣. منكر (denier). For example , ان علي خرج حقا / Ali definitely did leave

In this category, certainty markers are used to confirm and strengthen the informative meaning. The use of such markers, according to Al-Sakkaki's term, is known as "اعتبارات خطابية" (discourse variables) (Al-Sakaki, ١٩٨٧: ١٧١). The frequent use of certainty markers can be distinguished by two main categories: grammatical items that include قد، إنّ، أنّ، لام التوكيد، نون التوكيد (really, truly, definitely) and lexical items that include (inclusion, oath, cognate object, emphatic adverbials), and أفعال اليقين (certainty verbs) such as أعلم، أجزم، أعتقد (believe, assure, and know) (Fathi, ٢٠١٩: ٢٥). Consider the following example.

٣-أنت **_الشك_** راحل عنا غدا. فاذا وصلت الى حيث تقصد، فاذكرنا بالخير وال تقس في حكمك علينا. (Saleh, ١٩٩٧: ٥٣)

- Tomorrow, **_without doubt_**, you will be leaving us. When you arrive at your destination, think well of us and judge us not too harshly.

(Davies, ١٩٦٧:٩٤)

The writer used the lexical item ()الشك for assertion . .

According to Eskander (٢٠٠٣), emphatic circumlocution is subdivided into three categories: التكرار (repetition), الايغال(hyperbole) and التذييل (tagging).

### ٢,٢,١. (Repetition)التكرار

It is one of the most important rhetorical categories. It is defined as repeating a lexical item or a phrase to show emphasis. Thus, the repetition is used for a purpose; otherwise, it is considered اطالة(Redundancy). This type of circumlocution depends on the context and is used for many purposes such as assertion , comprehension , attracting the addressee to the discourse , repetition ,warning and reproach (Al-Hashimi, ١٩٩٩ :٢٠٣ ; Abbas , ١٩٩٧: ٤٨٩). One should differentiate between repetition and redundancy. Repetition can be a good category to use in writing since it adds emphasis to the meaning while redundancy cannot make a sense since it does not add anything to the meaning i.e. the repetition of a word does not add anything to the previous usage; it just repeats what has already been said. Consider the following example to clarify the function of this subcategory.

٤.  *يا بُني* أنك لن تمكث ...
    ليس هذا *يا بُني* غباراً ...
    هاك *يا بُني* هذا الشبكة من "التل" ...

(Saleh, ١٩٩٧:٣٣)
- ***My son***, that you would not stay long…,
- This , ***my son*** ,would not be dust …,
- Take this gauze netting , ***my son***, …

(Davies, ١٩٦٧:٨٣)

The writer repeated (يا بني) more than ones time in the same paragraph and in different sentences to persuade and to accept advice.

### ٢,٢,٢ (Hyperbole) الإيغال

It is an exaggeration of description and representation. It is used mostly in poetry. الإيغال (Hyperbole) can be distinguished from التتميم(Completion) in that the omission of the former does not affect the meaning while the omission of the latter affects the meaning (Ibn Abi I-Asba`, ١٩٥٧: ٩٢ ; Al-Sutyuti , ١٩٨٨: ٢٧٨). Consider the following example:

٥.  وصوته إذا تكلم يخرج مبحوحاً مكتوماً كصوت الوابور **إذا انكتم نفسه وشحر**. (Idris, ٢٠١٩:٣٣)
- His voice was hoarse and loud, like ***a rusty*** steam engine.

(Husni&Newman, ٢٠٠٨:٢٧٢)

The writer used the phrase (اذا انكتم نفسه و شحر ) for more assertion and clarification.

### ٢,٢,٣ (Tagging)التذييل

The speaker can tag on his discourse after completing the meanings in the utterances and its purpose is to ascertain the discourse (Ibn Abi I-Asba`, ١٩٥٧: ١٥٥). It is used to make sure that the audience successfully follows the flow of discourse (Fathi, ٢٠٠٥: ٩٠). Consider the following example:

٦.  والسرعة غير مطلوبة ابداً، *والعجلة من الشيطان*...
(Idris, ٢٠١٩:٢٦٩)
- And there is never any need for speed or haste .***As the saying goes:"The devil takes a hand in what is done in haste".***

(Husni&Newman, ٢٠٠٨:٢٦٨)

The writer used the Arabic proverb (العجلة من الشيطان ) to emphasize the meaning of the text preceded it.

### ٢,٣. (Directive circumlocution) الإطناب التوجيهي

It means directing the context to the intended meaning by using certain pragmatic elements to clarify and identify the meaning to the addressees and avoid misunderstanding .In this sense, it is used to make an interaction between the addressee and the text. It is categorized under the following headings:

### ٢,٣,١. ( Bracketing )الإعتراض

According to Ibn AL-Atheer (١٩٩٠: ١٧٢), الإعتراض(bracketing) is لفظ فيه أدخل كلام "كل , مفرد او مركب لو سقط لبقي الأول على حاله" ,'an utterance which is introduced into a single or compound expression. If it is omitted, the meaning will not change'. The purposes of الإعتراض (bracketing) , are التعظيم(deification) , التحسر(sigh), زيادة التوكيد ( to increase confirmation) , التنبيه ( warning ) and دعاء (supplication) (Ateeq, 2009 : 194). Consider the following example:

7. أهل البندر ال ينامون ال في اخريات الليل – *ذلك ما أعلمه عنهم.*

(Saleh, 1997:46)

- Townsfolk do not go to sleep till late at night – ***I know that of them***.

(Davies, 1967: 90)

The attitude of the writer towards the proposition addressed is clear in the text mentioned above. He used (ذلك ما أعلمه عنهم) as a commentary item to comment on the proposition addressed.

### 2.3.2. الاحتراس (Hedging)

According to Abbas (1997: 495), الاحتراس ( hedging ) is "المحافظة على المعنى من كل ما يفسده", "Preserving the meaning from anything that disturbs and changes it ". It is used when the speaker withholds commitment to the statement in such a way that he can soften the speech (Al-Hashimi, 1999:205). It can be realized by hedging devices such as قد، ربما، بعض (may be, perhaps, some) and أفعال الظن (uncertainty verbs) such as حسب، ظن (think, fancy) which show that the addresser is not sure about the truth-value of the propositions. Such devices may express interpersonal meaning to show the addresser's attitude towards the content of the message and the addressee. The main purpose of this subcategory is to soften the discourse and show doubtfulness. Consider the following example:

8. *اغلب الظن* يا بني أنها نمت وحدها، ...

(Saleh, 1997: 38)

- ***Most probably***, it grew up by itself,

(Davies, 1967:86)

The writer used the hedging device (اغلب الظن) to show his attitude towards the content of the text .

### 2.3.3. التتميم (Completion )

It is the addition of one or more than one word used for aesthetic values. If they are omitted, the speech will be prosaic (Al-Hashimi, 1999:205). Consider the Following example:

9. العربات الجميلة*المريحة.*

(Saleh ,1997: 34)

- ***The fine*** comfortable buses .

(Davies, 1967:83)

The original writer used the word ( المريحة) in the text to complete the description of (العربات).

According to what mentioned previously ,the following table shows the types of circumlocution in Arabic.

| الاطناب (circumlocution) | الاطناب البياني (Eloquence circumlocution) | الايضاح بعد الابهام (Illustration after Amphibology) |
|---|---|---|
| | | ذكر الخاص بعد العام (Mentioning the specific item after the general one) |
| | الاطناب التوكيدي (Emphatic circumlocution) | التكرار(Repetition) |
| | | الايغال(Hyperbole) |
| | | التذييل(Tagging) |
| | الاطناب التوجيهي (Directive circumlocution) | الاعتراض(Bracketing ) |
| | | الاحتراس(Hedging) |
| | | التتميم(Completion) |

*Table (1) : Classification of Arabic Circumlocution*

## ٣- The concept and the categories of English Metadiscourse :

Harris (١٩٧٠) to refer to discourse about discourse first introduced the term "metadiscourse". In his argument, he states the following:

*"The various sentences of a text differ in informational status, and even certain sentences which may be of interest to readers of the text may not be requested or useful in retrievals. These are metadiscourse kernels which talk about the main material". (Ibid: 466)*

Depending on the Hallidayian school of language ,researchers and scholars ( see, Williams, ١٩٨١; Vande Kopple, ١٩٨٥; Crismore et al., ١٩٩٣ (distinguish between textual metadiscourse and interpersonal metadiscourse According to Crismore et al. (١٩٨٣ : ١٢ ) , all informative texts have a propositional content level (the ideational part ) called the primary discourse and other texts have another level, the contentless level called metadiscourse. As a central pragmatic construct, metadiscourse helps text producers project themselves into text, arrange and organize the content to influence readers' understanding of both the text and their attitude towards its content and the audience (Hyland, ١٩٩٨: ٤٣٧).It includes comments about the discourse plans, the author's attitudes, the author's confidence in his following assertion, and the use of self-references and references to the readers i.e. the interpersonal part .Regarding the taxonomies of metadiscourse , different classifications are proposed , most of them sharing a functional Hallidayian approach in that are divided into two main categories: textual and interpersonal. Scholars and researchers such as (Crismore et al., ١٩٩٣Williams, ١٩٨١; Vande Kopple, ١٩٨٥ ;) have, though different terms used, agreed that categories and subcategories could be illustrated in the following table:

| | | Item | | Example |
|---|---|---|---|---|
| Metadiscourse in English | TEXTUAL METADISCOURSE | Textual Connectives | Additives | Also, furthermore, in addition to, |
| | | | Adversatives | However ,but ,nonetheless |
| | | | Temporal | Now, later, then |
| | | | Sequencers | First, second |
| | | | Causal | Therefore, thus, so |
| | | Code Glosses | | by this I mean |
| | | Illocution Markers | | I state again that |
| | | Reminders | | As I mentioned earlier |
| | | Narrators | | According to X |
| | INTERPERSONAL METADISCOURSE | Hedges | | may, perhaps |
| | | Certainty Markers | | certainly, really ,indeed |
| | | Attitude Markers | | surprisingly, doubtfully |
| | | Commentary | | You may not agree that |

**Table (٢): Categories and Subcategories of Metadiscourse**

## ٤- The model adopted in this study.

One of the most interesting discussion of the notion of equivalence can be found in Nida's (١٩٦٤) model. According to him , the translator faces difficulties in the process of translation especially that of differences in linguistic system and culture between (SL) and (TL) and there is no definition of translation can avoid some difficulties such as translating literary texts .In these texts, there are always conflicts between form and content and the using of formal and dynamic equivalents (ibid. : ١٦١) . He (١٩٦٤: ١٥٦) mentions that the variety of translations depends on three factors, the nature of the message, the purposes of the author and the translator, and the type of the audience.

He sets two basic orientation in translation, in which the translator seeks to reach the closest equivalent, a translation oriented towards formal equivalence and a translation oriented towards dynamic equivalence. According to Nida (١٩٦٤ : ١٥٩), there are two types of translation equivalence : formal equivalence and dynamic equivalence .Formal equivalence is used to reproduce some formal elements which include grammatical units which may include translating noun to noun and consistency in word usage ( Nida ,١٩٦٤ : ١٦٥) .Formal equivalence translation includes the message itself with the form and content such as rendering poetry to poetry , sentence to sentence and concept to concept (ibid:١٥٩). Dynamic equivalence,on the other hand, is based on the equivalent effect in which it reproduces , by using the receptor language , the closest equivalent of the message of the source language. It is a receptor oriented which is unlike formal equivalence of a source message oriented. For Nida (١٩٦٤ : ١٥٩), Dynamic Equivalence is achievable when the relationship between the message and receptor of (TT) is the same as the relationship between the message and receptor of (ST).. Based on Dynamic equivalence , Nida ( ١٩٦٤ : ٢٢٦ ) suggests three techniques and adjustments in the process of translation : additions , subtractions and alterations . Additions are used for some cases in which the they are necessary such as explain elliptic expressions , remove the ambiguity of the lexical item in the TL to avoid misleading reference ,the change of the linguistic category when necessary and add connectors when required as in the case of translating from English into Arabic. Subtractions are used when there is unnecessary repetition, specified references, conjunctions and adverbs. Alterations are the changes used due to the differences between the (SL) and (TL) and semantic problems. There are three main types of these changes: ١) Changes due to problems caused by transliteration ٢) Changes due to structural differences between (SL) and (TL) ٣) Changes due to semantic problems , especially with idiomatic expressions.

### ٥- Data Collection and Analysis Procedure

The data of this study draws on four Arabic short stories entitled (Doumat Wad Hamid ) written by Tabe Saleh (١٩٩٧), a well-known Sudanese writer ,and translated into English by Denys Jobnson-Davies (١٩٦٧) ; ( A Hand in the Grave ) written by Ghassan Kanafani (٢٠١٥), a well-known Palestinian writer, and translated into English by Hilary Kilpatrick (١٩٧٨) ; ( A Tray from Heaven) written by Yusuf Idris (٢٠١٩), a well-known Egyptian writer, and translated into English by Ronak Husni and Daniel L. Newman (٢٠٠٨) ; and ( The Picture ) written by Latifa Al-Zayyat (١٩٨٦), a well-known Egyptian writer, and translated into English by Dalya Cohen-Mor (٢٠٠٥). One of the main procedure in text analysis is to set a procedure for segmenting the text into parts . The most important criterion of this study is to decide whether the linguistic expression is considered as metadiscourse or a prepositional content. The decision of the analysis is based on the function of the item in a particular context ; thus , the analysis will be functional analysis rather than a linguistic . Both languages , Arabic and English , use individual words ,phrases and whole clauses in realizing metadiscourse function . What is still questionable is the presence and the absence of metadiscourse items. For this reason , it is suggested to use the method of dealing with any linguistic material as theme or the topic, termed in Arabic as ( المسند اليه )and the rheme or predicate , termed in Arabic as (المسند) (Al-Hashimi, ١٩٩٩ : ١٠١,١٣١).Thus ,the function of the topic theme ( المسند اليه) is "announcing the topic rather than offering new information about the chosen subject matter" (Lautamatti, ١٩٧٨: ٧٢) and the function of the rheme or predicate )المسند( is adding new information about the theme . When theme and rheme are identified , the identification of Lautamatti's (١٩٧٨) topical subjects and non- topical subjects i.e. metadiscourse in the texts becomes easier .

Depending on Lautamatti (١٩٧٨) procedure, the following Arabic examples and their translations which are taken from the samples of the study will illustrate the identification of circumlocution , in which it is supposed to be equivalent to metadiscourse, and ( المسند اليه theme) and المسند (rheme):

١٠. لعل الباخرة حينئذ تقف عندنا .

(Saleh ,١٩٩٧ : ٥٣)

- ***Maybe*** then ***the steamer will stop*** at our village .

(Davies ,١٩٦٧:٩٤)

In the above text , لعل (maybe ) is the non-topical discourse (metadiscourse) . Its function is الاحتراس (hedging) . The topical discourse is segmented into two parts الباخرة (the steamer ( is considered as مسند اليه (theme) and تقف ( will stop ) is considered as مسند(rheme) .

Circumlocution categories and subcategories used in the data are classified according to their functions . Thus, circumlocution categories are classified into textual metadiscourse which helps the reader to understand the propositional meaning  and interpersonal metadiscourse which decodes the text producer's attitude .

### ٥.١. Analysis of Interpersonal Metadiscourse .

By using interpersonal metadiscourse which decodes the text producer's attitude ,interpersonal categories and subcategories of circumlocution will be analyzed and discussed .

#### ٥.١.١  (Emphatic circumlocution .) االطناب التوكيدي

This category is considered one of the most important categories of  circumlocution in which it takes into consideration the state of the audience in a particular context whether he was خااالاي الاذهان (open-minded) ماتاردد (uncertain), or ماناكار (denier) .Certainty markers are used to confirm and strengthen the informative meaning (Al-Sakaki ,١٩٨٧: ١٧١).

Most problematic areas are found in the renditions of the category . Translators could not manage to render this category appropriately. In so doing , different techniques are adopted by the translators for rendering this category . In the following source text ,the items of circumlocution are three certainty markers (بل) , ( الم التوكيد) and (ان).They are used by the writer for expressing high level of certainty .

١١.      أينما كنت في هذه البلدة تراها ... بل انك لتراها ...و أنت في رابع بلدة من هنا (٣٨ : ١٩٩٧ , (Saleh

- Wherever you happen to be in the village you can see it ; ***in fact*** ,You ***can even*** see it from four village a way .

(Davies ,١٩٦٧: ٨٥)

In so doing , the translator inappropriately managed his rendition of certainty. He did not grasp the function of certainty because he failed to render the certainty of the message used by the original writer . He opted for dynamic equivalence using the technique of semantic alteration .Considering the co-text and the context of the underlined items mentioned above , (بل) is a certainty marker whereas the translator neutralized it in his rendition into (in fact) .Also , he rendered ان into (can ) and الم التوكيد into (even ) .In this sense , the translator depleted the (ST) and neutralized it . The appropriate rendition could be appropriate by opting for formal equivalence and using items that have high level of certainty . Consider the following suggested rendition:

- Wherever you happen to be in the village you can see it ; ___*indeed*___ ,You **will certainly** see it from four village a way .

According to Eskander (٢٠٠٣) ,emphatic circumlocution is subdivided into three categories:

٥,١,١,١ ( Repetition ) التكرار
This rhetorical category is used frequently in the short stories under the study . As it is mentioned above , this type of circumlocution depends on the context and is used for many purposes such as assertion , comprehension , attracting the addressee to the discourse , repetition ,warning and reproach (Al-Hashimi, ١٩٩٩ :٢٠٣ ; Abbas , ١٩٩٧: ٤٨٩).

In some cases , translators unsuccessfully and inappropriately rendered the function of repetition . The following text shows that the (SL) writer repeated the lexical item (الراحة) to emphasize that (الراحة) is an ugly thing .

١٢.    لكن *الراحة* كلمة بشاعة عند الفالحين . *الراحة* اهانة لخشاونتهم وقدرتهم الخارقة على العمل التي ال و تكل .

(Idris,٢٠١٩:٣٢)

- The word "___*rest*___" was considered ugly among the farmers **,** ___*as well as*___ an insult to their toughness and to their extraordinary ability to work indefatigably .

(Husni&Newman, ٢٠٠٨:٢٧٠)

In his rendition, the translator used semantic alteration , he inappropriately opted for dynamic equivalence. The appropriate rendition could be achieved by following formal equivalence and repeating the word ( rest ) . Consider the following suggested rendition .

- The word "rest" was considered ugly among the farmers **.** ___*Rest*___ is an insult to their toughness and to their extraordinary ability to work indefatigably .

٥,١,١,٢.(Hyperbole)االيغال

Rarely , hyperbole is found in novels and short stories since it is much used in poetry . Hyperbole is an exaggeration of description and representation and if it is omitted ,it will not affect the meaning . It is divided into two types: ايغال تخيير(Optional Hyperbole) and ايغال احتياط (Spare Hyperbole).

In the following text , it is a clear that the (SL) writer used the phrase ( شحر اذا انكتم نفساه و ) for more assertion and clarification .

١٣.    وصوته اذا تكلم يخرج مبحوحاً مكتوماً كصوت الوابور **اذا انكتم نفسه وشحر** . (Idris, ٢٠١٩:٣٣)
- His voice was hoarse and loud , like ___*a rusty*___ steam engine .
(Husni&Newman, ٢٠٠٨
:٢٧٢)

In his rendition , the translator was unsuccessful in rendering the function of this subcategory. He opted for dynamic equivalence using semantic alteration . He rendered the phrase ( اذا انكتم نفساه و شاحر) into (rusty) . To achieve appropriate rendition , formal equivalence should be used . Consider the following suggested rendition .

- His voice was hoarse and loud , like a steam engine ___*when it is muffled and blackened with soot.*___

### 3.1.1.5. (Tagging) التذييل

The speaker can tag on his discourse after completing the meanings in the utterances and its purpose is to ascertain the discourse ( Ibn Abi I-Asba` , ١٩٥٧: ١٥٥). There are two types of Tagging , one of them does not add anything to the first meaning but confirms it and the other is used as a proverb that could be mentioned independently(ibid).

In some cases , the translators opted for alteration technique . In the following text , the original writer used (بقية الدوم بالنسبة اليها كقطيع الماعز بينهن بعير) to emphasize the meaning of the text preceded it.

١٤.        ثم صااااعاد تل ، فلما بل قمتاه رأك غابة كلاة من الدوم في وسااااطهاا دومة – دومة طويلاة، *بقيَّةُ الدوم أيِّ اليها كقطيع الماعز بينهن بعير* .

(Saleh ,١٩٩٧ : ٣٩)

- how he climbed a hill and on reaching the top espied a dense forest of doum trees with a single tall tree in the center ***which in comparison with the others looked like a camel amid a herd of goats***;

(Davies ,١٩٦٧: ٨٦)

In his rendition ,the translator rendered it successfully and appropriately .He opted for dynamic equivalence and using the technique of semantic alteration . He rendered the (ST) ( بقية الدوم بالنسبة اليها كقطيع الماعز بينهن بعير) into (TT) (which in comparison with the others looked like a camel amid a herd of goats ) .

### 5.1.2 (Directive circumlocution) الاطناب التوجيهي

It means directing the context to the intended meaning by using certain pragmatic elements to clarify and identify the meaning to the addressees and avoid misunderstanding .In this sense , it is used to make an interaction between the addressee and the text. It is categorized under the following headings :

### 5.1.2.1. ( Bracketing )الاعتراض

It is an utterance which is introduced into a single or compound expression. If it is omitted , the meaning will not change (Ibn AL-Atheer ,١٩٩٠: ١٧٢) .The purposes of الاعتراض , are ( warning ) التنبيه , (Increase confirmation) زيادة التوكيد , (sigh)التحسر , (deification)التعظيم , دعاء (Supplication) (Ateeq, ٢٠٠٩ : ١٩٤ ) .

One of the most problematic areas in rendering this device is that the translators could not differentiate between circumlocution items and propositional ones as is the case in the following text.

١٥.        هكذا يبدو في كل صابا ، أنه ال يفعل شايئا" ساوك ان يفت – *طوال ما قبل الفطور* – عن سابب يلقي عليه ثقل غضبه.

(Kanafani ,٢٠١٥:٦٠)

- That is how he appeared every morning ,doing nothing ***the whole time before breakfast*** except search for a pretext to unburden himself of his rage ;

(Kilptrick ,١٩٧٨:٦٩)

It is clear that the original writer used (طوال ما قبل الفطور) as bracketing for the purpose of assertion. However, the translator inappropriately rendered it. He could not differentiate between the circumlocution items and the propositional material. In his rendition , the translator mixed bracketing in the (ST) phrase (طوال ما قبل الفطور) with the proposition text preceded it. Being so, he could not grasp the function of this subcategory .He opted for dynamic equivalence using structural alteration. To achieve the appropriate rendition , formal equivalence should be used. The bracketing phrase should be rendered as metadiscourse preceded and followed by commas . Consider the following suggested rendition :

- That is how he appeared every morning , ***he didn't do anything*** , ***the whole time before breakfast*** , except search for a pretext to unburden himself of his rage ;

### ٥.١.٢.٢ الاحتراس (Hedging)

This subcategory of directive circumlocution is used frequently in the short stories under the study . It is "المحافظة على المعنى من كل ما يفساده ويغيره" , "Preserve the meaning from everything that corrupts and changes it " ) Abbas , ١٩٩٧: ٤٩٥) . It is used when the speaker withholds commitment to the statement in such a way that he can soften the speech (Al-Hashimi, ١٩٩٩ :٢٠٥).

One of the most important problematic areas in the rendition of this category is that translators are confused between the categories of certainty and hedging . The following text is an instance of this problem in which the translator rendered what is a hedging marker into a certainty one .

١٦.  *وكادوا* يفتكون به ، لولا أني تدخلت فأنتزعته من براثنهم،...

(Saleh ,١٩٩٧: ٤٣)

- And ***would certainly*** have killed him if I had not intervened and snatched him from their clutches .

(Davies, ١٩٦٧:٨٨)

The original writer used the item (كاادوا) which associated with (لوال) to express the function of hedging .In his rendition , the translator unsuccessfully and inappropriately rendered this subcategory. He opted for dynamic equivalence using semantic alteration technique . He rendered hedging to high degree of certainty. The appropriate rendition could be achieved by following the same orientation of equivalence using grammatical alteration . Consider the following suggested rendition :

- And they were ***about to*** kill him if I had not intervened and snatched him from their clutches.

### ٥.١.٢.٣ التتميم (Completion)

It is the addition of one or more than one word used for aesthetic values. If they are omitted, the speech will be prosaic (Al-Hashimi, ١٩٩٩ :٢٠٥). There are three types of completion ( تتميم احتياط Exaggeration Completion ) and تتميم مبالغة ( Reduction Completion ), تتميم نقص Spare Completion ) (Ibn Abi I-Asba` , ١٩٥٧: ٤٦). Consider the following example:-

Translators of the short stories under the study successfully and appropriately rendered this subcategory . Consider the following text .

١٧.    العربات الجميلة *المريحة*.

(Saleh ,١٩٩٧: ٣٤)

- ***The fine*** comfortable buses .

(Davies, ١٩٦٧:٨٣)

The original writer used the word (المريحة) in the text to complete the description of (الاعارباات). The translator grasped the function of this subcategory . He opted for dynamic equivalence using word order alteration . In (TT) , adjectives tend to be pronominal : Adj + N

, while in (ST) they are post nominal . In his rendition , the translator successfully and appropriately managed to render it .

### ٥.٢ Analysis of Textual Metadiscourse

By using textual metadiscourse which is used to facilitate the understanding of the propositional meaning, the main textual categories and subcategories of الإطناب (circumlocution) in (STs) and their renditions in the (TTs) will be analyzed and discussed.

#### ٥.٢.١ (Eloquence ircumlocution)الإطناب البياني

Eloquence circumlocution as is discussed earlier , it is used to reveal and clarify the meaning to access its variant purposes . It is subdivided into two categories ذكر الخاص بعد العام (Mentioning the specific item after the general one) and الإيضاح بعد الإبهام ( illustration after amphibology). Each subcategory of this type of circumlocution with examples from the short stories under the study in (STs) and their renditions into (TTs) will be analyzed and discussed to see whether the translators have successfully and appropriately conveyed them to (TTs).

##### ٥.٢.١.١ ذكر الخاص بعد العام (Mentioning the specific item after the general one).

It means paying attention to the importance of the specific item that is mentioned after the general one . Usually , حروف العطف (coordinating particles) are used in this type of circumlocution ( Abbas , ١٩٩٧: ٤٨٦ ).

One of the problematic renditions of this subcategory is illustrated in the following example . It is clear that the original writer used (منية النصر Munyat al-Nasr) specifically before الواسعة كغيرها من بلاد هلال(Like other towns ) to highlight the importance of this city.

١٨. _النصر كغيرها من بلاد هلال / الواسعة_ تتشاءم من يوم الجمعة ، واي حادث يقع فيه لابد انه _وفي_ كارثة اكيدة ...

(Idris, ٢٠١٩:٣٢)

- **_Like other towns_** , **_Munyat al-Nasr_** was superstitious about Friday , and any event that took place on that day was viewed as a sure catastrophe.
(Husni&Newman, ٢٠٠٨:٢٧٠)

In his rendition , the translator has unsuccessfully and inappropriately managed to render this subcategory . He opted for dynamic equivalence . He used structural alteration technique since he did not mention منية النصر(Munyat al-Nasr) in the first position before كغيرها من بلاد هلال الواسعة (like other towns ) . So, it can be seen that the translator did not capture the function of this subcategory . The appropriate rendition could be achieved by opting for formal equivalence . Consider the following suggested rendition.
- **_Munyat al-Nasr_** , **_Like other towns_** , was superstitious about Friday , and any event that took place on that day was viewed as a sure catastrophe.

##### ٥.٢.١.٢ ( illustration after amphibology ) الإيضاح بعد الإبهام

It means that the speaker mysteriously mentions the meaning, then he clarifies it during his speech . Here , the meaning is presented in two different ways: one of them is الإبهام (amphibology ) and the other is الإيضاح (Illustration) . This subcategory is used to give the reader a sense of suspense (Al-Sutyuti , ١٩٨٨: ٢٧٢ ).

٢٥٢

After examining this subcategory and its rendition in the short stories under the study, the translators succeeded in achieving appropriate renditions of this subcategory. Consider the following examples in which the writer clarified to the reader what really meant by (ذباب البقر) since if we stopped till (ذباب البقر), the sentence would be ambiguous:

19. صيفاً فتجد عندنا ذباب البقر – *ذباب ضخم كحملان الخريف*، كما نقول بلهجتنا. (33 : 1997, Saleh) و

- If you were to come to us in summer you would find the horse flies with us – ***enormous flies the size of young sheep*** , as we say .

(Davies, 1967:83)

As the function of this subcategory in Arabic is defining and clarifying , it shares the same function of what is termed 'code glosses' in English textual metdiscourse .In this sense ,the rendition was appropriate since the translator preserved the function of this category by opting for the formal equivalence .

**The following table will illustrate in short the analysis discussed previously :**

| No. | SL item | TL item | Type of Equivalence | App. |
|---|---|---|---|---|
| 1 | بل و ان و الم التوكيد | in fact , can , even | Dynamic | - |
| 2 | الراحة | As well as | Dynamic | - |
| 3 | اذا انكتم نفسه وشحر | A rusty | Dynamic | - |
| 4 | بقياة الادوم بالنساااباة اليهاا كقطيع الماعز بينهن بعير | which in comparison with the others looked like a camel amid a herd of goats | Dynamic | + |
| 5 | طوال ما قبل الفطور | the whole time before breakfast | Dynamic | - |
| 6 | كادوا | would certainly | Dynamic | - |
| 7 | المريحة | The fine | Dynamic | + |
| 8 | منياة النصااااار ، كغيرهاا من بلاد هلال الواسعة | like other towns | Dynamic | - |
| 9 | ذباب ضخم كحملان الخريف | enormous flies the size of young sheep | Formal | + |

Table (3) : Analysis of (SL) and (TL) items

### 6- Conclusions

Based on a linguistic and functional analysis of the data under study ,several points are concluded. The data under the study shows that circumlocution exists widely in Arabic short stories with its different categories .Interpersonal metadiscourse categories and subcategories such as certainty ,repetition and hedging used more frequently than others .Comparing circumlocution categories and subcategories used in (STs) and their renditions into the (TTs) , it is founded that there are lots of similarities between the categories of circumlocution and metadiscourse ones. These similarities are clear when appropriate renditions achieved . However , the analysis revealed that the translators of the short stories under the study had several problematic issues that led to inappropriate renditions of these categories including :

1. The translators confused between the categories and subcategories of circumlocution such as rendering what is hedging marker to certainty marker (see 5,1,2,2, example 16 ).

2. They could not make a distinction between circumlocution and propositional content (see ,5,1,2,1.,example 15).

٣. They subtracted circumlocution items used in the (SL) texts .(see ٥,١,١,١,example ٥ )

It can be concluded that interpersonal metadiscourse categories which are الطناب التوكيدي (Emphatic circumlocution) and الطاناااب الاتااوجاياهاي (Directive circumlocution) are the most problematic areas that the translators faced in the process of translation .These two categories are equivalent to Hedging and Certainty markers in English interpersonal metdiscourse .However , the translators in some areas captured the function of these categories and in other are failed .This study is a call to pay much attention to the study of circumlocution in Arabic and its use in different genres. However, applying circumlocution categories to the textual and interpersonal categories of Metadiscourse, provides a more systematic system of classifying circumlocution in Arabic . This tentative attempt shows that Arabic uses metadiscourse as much as English does, and consequently it is a universal phenomenon . Circumlocution and metadiscourse are a crucial feature in any discourse . This study also shows that circumlocution is not all of metadiscourse ,but it is an essential part of it. Thus , it is just a matter of using two different terms but the function is the same .